\documentclass{article} 
\usepackage{iclr2017_conference}
\usepackage{hyperref}

\usepackage{times}
\usepackage{epsfig}
\usepackage{amsmath}
\usepackage{amssymb}
\usepackage{url}
\usepackage{graphicx}
\usepackage{epstopdf}
\usepackage{wrapfig}
\usepackage{subfigure}
\usepackage{caption}

\title{Paying More Attention to Attention:\\ Improving the Performance of Convolutional Neural Networks via Attention Transfer}

\author{Sergey Zagoruyko, Nikos Komodakis\\
  Universit\'e Paris-Est, \'Ecole des Ponts ParisTech\\
  Paris, France\\
  \texttt{\{sergey.zagoruyko,nikos.komodakis\}@enpc.fr}
}

\iclrfinalcopy 

\begin{document}

\maketitle

\begin{abstract}
Attention plays a critical role in human visual experience.
Furthermore, it has recently been demonstrated that attention can also play an important role in the context of applying artificial neural networks to a variety of tasks from fields such as computer vision and NLP. In this work we show that, by properly defining attention for convolutional neural networks, we can actually use this type of information in order to significantly improve the performance of a student CNN network by forcing it to mimic the attention maps of a powerful teacher network. To that end, we propose several novel methods of transferring attention, showing consistent improvement across a variety of datasets and convolutional neural network architectures. Code and models for our experiments are available at \url{https://github.com/szagoruyko/attention-transfer}.
\end{abstract}

\section{Introduction}

As humans, we need to pay attention in order to be able to adequately perceive our surroundings. Attention is therefore a key aspect of our visual experience, and closely relates to perception - we need to keep attention to build a visual representation, possessing detail and coherence.

As artificial neural networks became more popular in fields such as computer vision and natural language processing  in the recent years, artificial attention mechanisms started to be developed as well. Artificial attention lets a system ``attend'' to an object to examine it with greater detail. It has also become a research tool for understanding mechanisms behind neural networks, similar to attention used in psychology.

One of the popular hypothesis there is that there are non-attentional and attentional perception processes. Non-attentional processes help to observe a scene in general and gather high-level information, which, when associated with other thinking processes, helps us  to control the attention processes and navigate to a certain part of the scene. This implies that different observers — with different knowledge, different goals, and therefore different attentional strategies — can literally see the same scene differently. This brings us to the main topic of this paper: how attention differs within artificial vision systems, and can we use attention information in order to improve the performance of convolutional neural networks ? More specifically, can a teacher network improve the performance of another  student network by providing to  it information about where it looks, i.e., about where it concentrates its attention into ?

\begin{figure}
  \vspace{-1.2cm}
  \centering
  \subfigure[] {
    \includegraphics[scale=0.6]{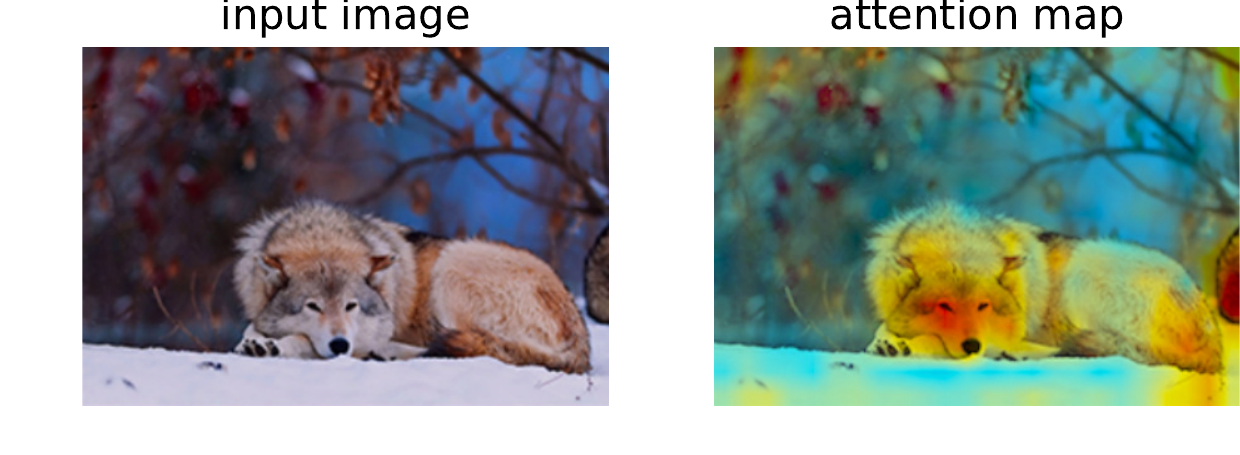}}
  \hspace{0.5cm}
  \subfigure[]{
    \includegraphics[scale=0.6]{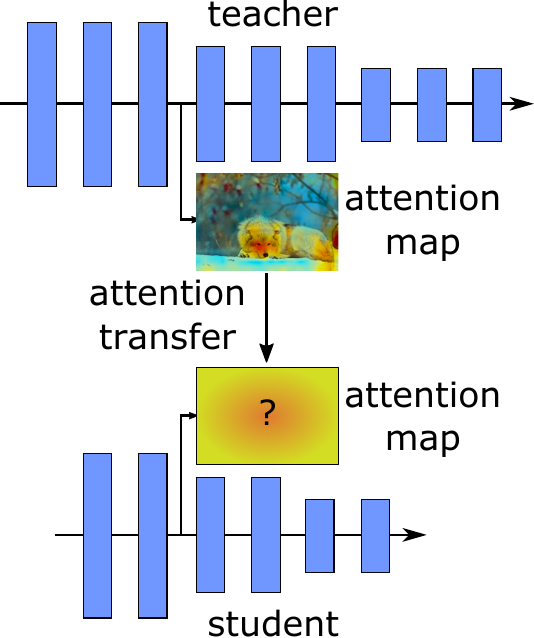}}
  \vspace{-0.3cm}
  \caption{\textbf{(a)} An input image and a corresponding spatial attention map of a convolutional network that shows where the network focuses in order to classify the given image. Surely, this type of map must  contain valuable information about the network.  The question that we pose in this paper is the following: can we use knowledge of this type to improve the training of CNN models ? \textbf{(b)} Schematic representation of attention transfer:   a student CNN is trained so as, not only to make good predictions, but  to also have similar spatial attention maps to those of an already trained teacher CNN. }
  \label{fig:wolf}
\end{figure}

To study these questions, one first needs to properly specify how attention is defined  w.r.t. a given convolutional neural network. To that end, here we consider attention as a set of \emph{spatial} maps
that essentially try to encode on which spatial areas of the input the network focuses most for taking its output decision (e.g., for classifying an image), where, furthermore, these maps can be defined w.r.t. various layers of the network so that they are able to capture both low-, mid-, and high-level representation information.
More specifically, in this work we define two types of spatial attention maps:  \emph{activation-based }and \emph{gradient-based}. We explore how both of these attention maps  change over various datasets and  architectures, and show that these  actually contain valuable information that can be used for significantly improving the performance of convolutional neural network architectures  (of various types and trained for various different tasks). To that end, we propose several novel ways of transferring attention from a powerful teacher network to a smaller student network with the goal of improving the performance of the latter (Fig.~\ref{fig:wolf}).

To summarize, the contributions of this work are as follows:
\begin{itemize}
  \item We propose attention as a mechanism of transferring knowledge from one network to another
  \item We propose the use of both activation-based and gradient-based spatial attention maps
  \item We show experimentally that our approach provides significant improvements across a variety of datasets and deep network architectures, including both residual and non-residual networks
  \item We show that activation-based attention transfer gives better improvements than full-activation transfer, and can be combined with knowledge distillation
\end{itemize}

The rest of the paper is structured as follows: we first describe related work in section \ref{sec:related_work}, we  explain our approach for  activation-based and gradient-based attention transfer in section \ref{sec:attention_transfer}, and then present experimental results for both methods in section \ref{sec:results}.
We conclude the paper in section \ref{sec:conclusions}.

\section{Related work}\label{sec:related_work}

Early work on attention based tracking \cite{NIPS2010_4089}, \cite{Denil2012a} was motivated by human attention mechanism theories \cite{rensink} and was done via Restricted Bolzmann Machines. It was recently adapted for neural machine translation with recurrent neural networks, e.g.\ \cite{DBLP:journals/corr/BahdanauCB14} as well as in several other NLP-related tasks. It was also exploited in computer-vision-related tasks such as image captioning \cite{showattendtell}, visual question answering   \cite{YangHGDS15}, as well as  in weakly-supervised object localization \cite{Oquab15} and classification \cite{NIPS2014_5542}, to mention a few characteristic examples. In all these tasks attention proved to be useful.

Visualizing attention maps in deep convolutional neural networks is an open problem.
The simplest gradient-based way of doing that is by computing a Jacobian of network output w.r.t. input (this leads to attention visualization that are not necessarily class-discriminative), as for example in \cite{simonyan14}. Another approach was proposed by \cite{zeiler14} that consists of attaching a network called ``deconvnet'' that shares weights with the  original network and is used  to project certain features onto the image plane. A number of methods was proposed to improve gradient-based attention as well, for example guided backpropagation \cite{SprDosBroRied15}, adding a change in \textit{ReLU} layers during calculation of gradient w.r.t.\ previous layer output. Attention maps obtained with guided backpropagation are non-class-discriminative too. Among existing   methods for visualizing attention, we should also mention class activation maps \cite{zhou2016cvpr}, which are based on removing top average-pooling layer and converting the linear classification layer into a convolutional layer, producing attention maps per each class. A method combining both guided backpropagation and CAM is Grad-CAM by \cite{gradcam}, adding image-level details to class-discriminative attention maps.

Knowledge distillation with neural networks was pioneered by \cite{KD, caruana2006}, which is a transfer learning method that aims to  improve the training of a student network by relying on knowledge borrowed from a powerful teacher network. Although in certain special cases shallow networks had been  shown to be able to approximate deeper ones without loss in accuracy \cite{caruana2014}, later work related to knowledge distillation was mostly based on the assumption that deeper networks always learn better representations. For example, FitNets \cite{Romero-et-al-TR2014} tried to learn a thin deep network using a shallow one with more parameters. The introduction of highway \cite{highway} and later residual networks \cite{he2015deep} allowed training very deep architectures with higher accuracy, and generality of these networks was experimentally showed over a large variety of datasets. Although the main motivation for residual networks was increasing depth, it was later shown by \cite{Zagoruyko2016WRN} that, after a certain depth, the improvements came mostly from increased capacity of the networks, i.e.\ number of parameters (for instance, a wider deep residual network with only 16 layers was shown that it could learn as good or better representations as very thin 1000 layer one, provided that they were using  comparable number of parameters).

Due to the above fact and due to that thin deep networks are less parallelizable than wider ones, we think that knowledge transfer needs to be revisited, and take an opposite to FitNets approach - we try to learn less deep student networks. Our attention maps used for transfer are similar to both gradient-based and activation-based  maps mentioned above, which play a role similar to ``hints'' in FitNets, although we don't introduce new weights.

\section{Attention transfer}\label{sec:attention_transfer}
In this section we explain the two methods that we use for defining the spatial attention maps of a convolutional neural network as well as how we transfer  attention information  from a teacher to a student network  in each case.

\subsection{Activation-based attention transfer}

Let us consider a CNN layer and its corresponding activation tensor  $A \in R^{C\times H\times W}$, which  consists of $C$ feature planes with spatial dimensions $H\times W$.
An activation-based  mapping function $\mathcal{F}$ (w.r.t. that layer) takes as input  the above 3D tensor $A$ and outputs a  spatial attention map, i.e., a flattened 2D tensor defined over the spatial dimensions, or

\begin{equation}
  \mathcal{F}: R^{C\times H\times W}\rightarrow R^{H\times W}\enspace.
\end{equation}

\begin{wrapfigure}{r}{0.25\textwidth}
  \centering
  \includegraphics[scale=0.9]{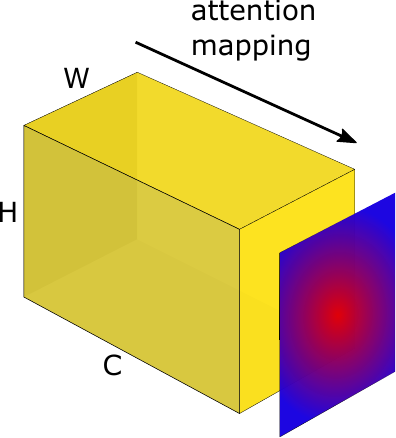}
  \caption{Attention mapping over feature dimension.}
  \label{fig:cube}
\end{wrapfigure}

To define such a spatial  attention mapping function, the implicit  assumption that we make in this section  is that the absolute value of a hidden  neuron activation  (that results when the network is evaluated on given input) can be used as  an indication about  the importance of that neuron w.r.t. the specific input.
By considering, therefore,
the absolute values of the elements of tensor $A$, we can construct a spatial  attention map by computing
 statistics of these values across the channel dimension (see Fig.~\ref{fig:cube}). More specifically, in this work we  will consider the following activation-based spatial attention maps:

\begin{figure}
  \vspace{-0.5cm}
  \centering
  \includegraphics[scale=0.6]{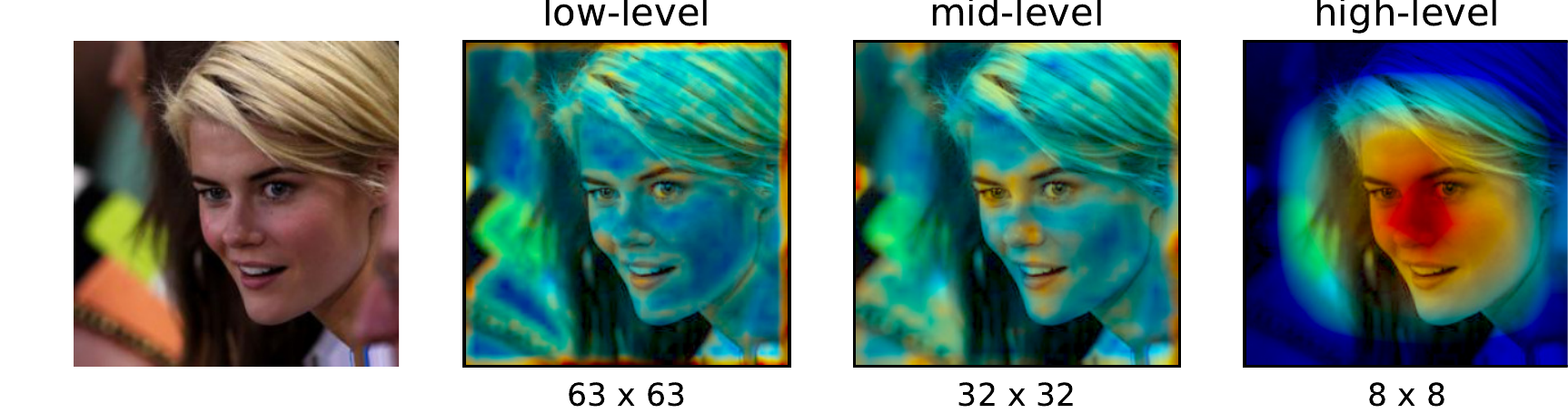}
  \caption{Sum of absolute values attention maps $F_{\mathrm{sum}}$ over different levels of a network trained for face recognition. Mid-level attention maps have higher activation level around eyes, nose and lips, high-level activations correspond to the whole face.}
  \label{figure:vggface_abssum}
\end{figure}

\begin{itemize}
  \item sum of absolute values: $F_{\mathrm{sum}}(A)=\sum^C_{i=1} |A_i|$
  \item sum of absolute values raised to the power of $p$ (where $p > 1$): $F^p_{\mathrm{sum}}(A)=\sum^C_{i=1} |A_i|^p$
  \item max of absolute values raised to the power of $p$ (where $p > 1$): $F^p_{\mathrm{max}}(A)=\max_{i=1,C}|A_i|^p$
\end{itemize}
where $A_i=A(i,:,:)$ (using Matlab notation), and max, power and absolute value operations are elementwise (e.g.\ $|A_i|^p$ is equivalent to \texttt{abs(A$_i$).$^{\land}$p} in Matlab notation).

\begin{figure}
  \hspace{-0.55cm}
  \centering
  \includegraphics[scale=0.38]{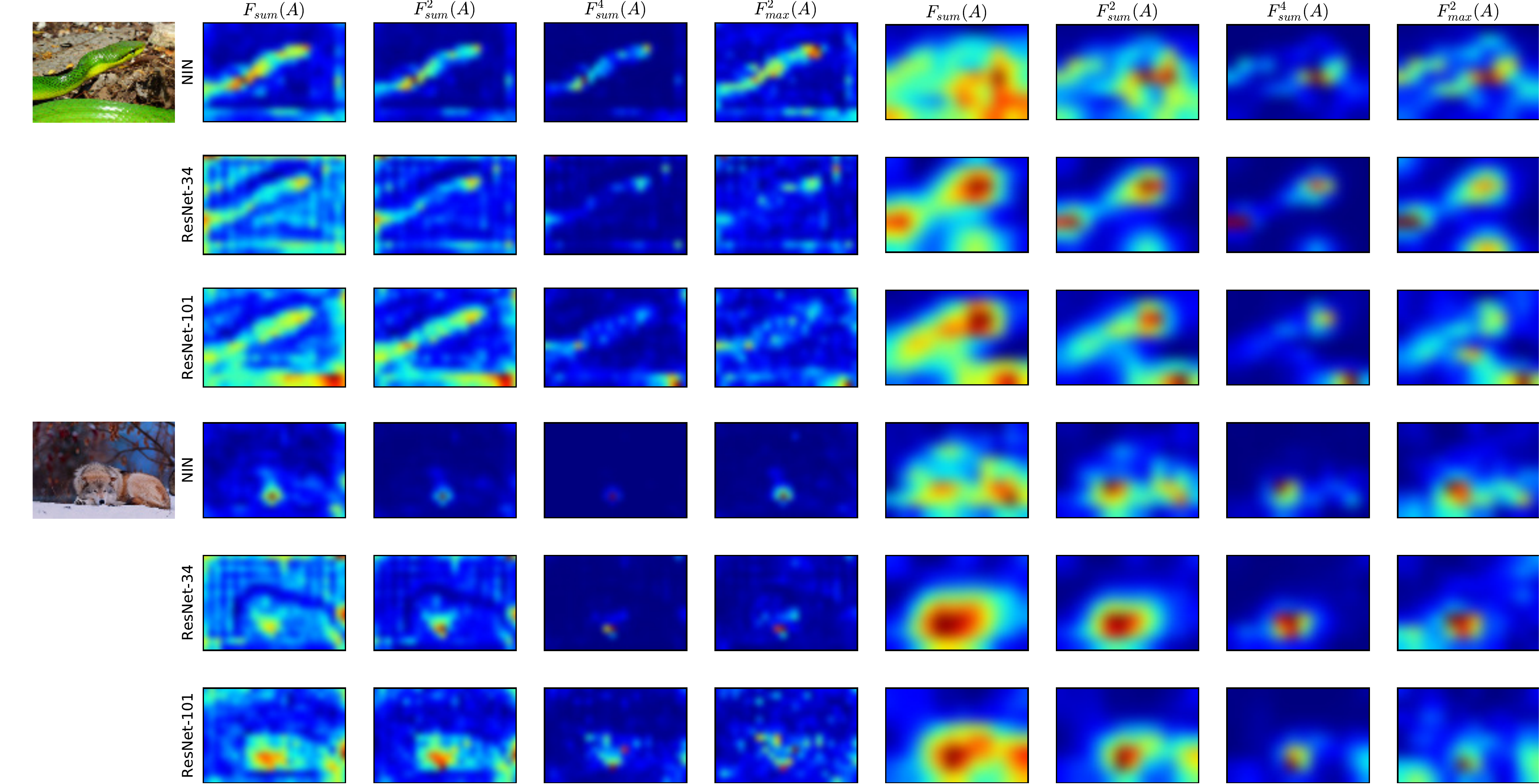}
  \caption{Activation attention maps for various ImageNet networks: Network-In-Network (62\% top-1 val accuracy), ResNet-34 (73\% top-1 val accuracy), ResNet-101 (77.3\% top-1 val accuracy). Left part: mid-level activations, right part: top-level pre-softmax acivations}
  \label{figure:imagenet_sumabs_maps}
\end{figure}

We visualized activations of various networks on several datasets, including ImageNet classification and localization, COCO object detection, face recognition, and fine-grained recognition. We were mostly focused on modern architectures without top dense linear layers, such as Network-In-Network, ResNet and Inception, which have streamlined convolutional structure. We also examined networks of the same architecture, width and depth, but trained with different frameworks with significant difference in performance. We found that the above statistics of hidden activations not only have spatial correlation with predicted objects on image level, but these correlations also tend to be higher in networks with higher accuracy, and stronger networks have peaks in attention where weak networks don't (e.g., see Fig.~\ref{figure:imagenet_sumabs_maps}).
Furthermore, attention maps focus on different parts for different layers in the network. In the first layers neurons activation level is high for low-level gradient points, in the middle it is higher for the most discriminative regions such as eyes or wheels, and in the top layers it reflects full objects. For example, mid-level attention maps of a network trained for face recognition \cite{Parkhi15} will have higher activations around eyes, nose and lips, and top level activation will correspond to full face (Fig.~\ref{figure:vggface_abssum}).

Concerning the different attention mapping functions defined above, these can have slightly different properties. E.g.:
\begin{itemize}
\item
Compared to $F_{\mathrm{sum}}(A)$, the spatial map $F^p_{\mathrm{sum}}(A)$  (where $p>1)$ puts more weight to spatial locations that correspond to   the  neurons with the highest activations, i.e., puts more weight to the most discriminative parts (the larger the $p$ the more focus is placed on those parts with  highest activations).

\item
Furthermore, among all neuron activations corresponding to the same spatial location, $F^p_{\mathrm{max}}(A)$ will consider only one of them to assign a weight to that spatial location (as opposed to $F^p_{\mathrm{sum}}(A)$ that will favor spatial locations that carry multiple neurons with high activations).
\end{itemize}

To further illustrate the differences of these functions we visualized attention maps of 3 networks with sufficient difference in classification performance: Network-In-Network (62\% top-1 val accuracy), ResNet-34 (73\% top-1 val accuracy) and ResNet-101 (77.3\% top-1 val accuracy). In each network we took last pre-downsampling activation maps, on the left for mid-level and on the right for top pre-average pooling activations in fig.~\ref{figure:imagenet_sumabs_maps}. Top-level maps are blurry because their original spatial resolution is $7\times7$. It is clear that most discriminative regions have higher activation levels, e.g.\ face of the wolf, and that shape details disappear as the parameter $p$ (used as  exponent) increases.

\begin{figure}[ht]
  \centering
  \includegraphics[scale=0.7]{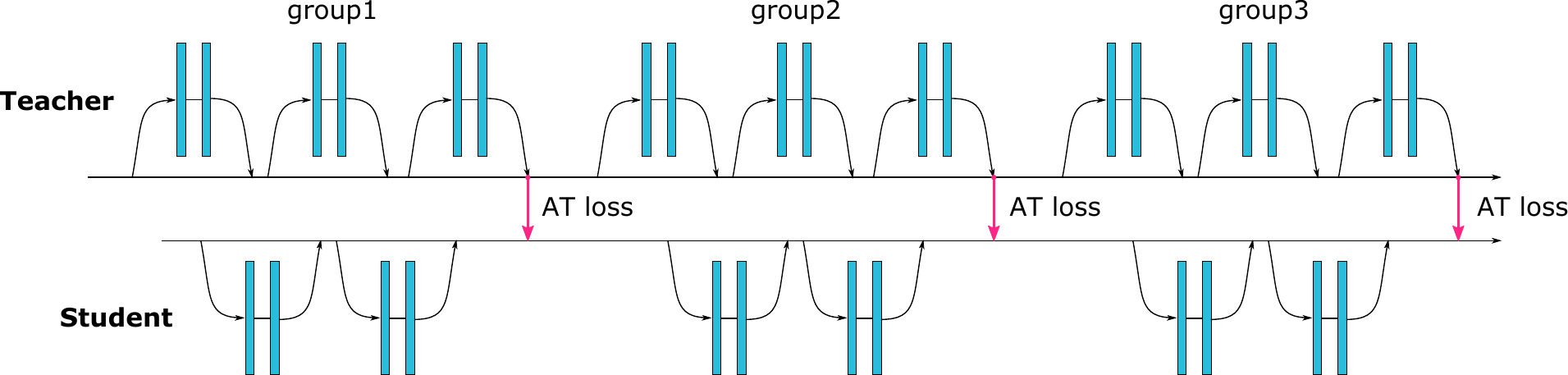}
  \caption{Schematics of teacher-student attention transfer for the case when both networks are residual, and the teacher is deeper.}
  \label{fig:resnet_attention_schematics}
\end{figure}

In attention transfer,   given the spatial attention maps of a teacher network (computed  using any of the above attention mapping functions), the goal  is to train a student network  that will not only make correct predictions but will also have   attentions maps that are similar to those of the teacher. In general, one can place  transfer losses w.r.t. attention maps computed across
several layers.
 For instance, in the case of ResNet architectures,
one can consider the following two cases, depending on the depth of teacher and student:

\begin{itemize}
  \item Same depth: possible to have attention transfer layer after every residual block  \item Different depth: have attention transfer on output activations of each group of residual blocks \end{itemize}

Similar cases apply also to other architectures (such as NIN, in which case a group refers to a block of a $3\times3$, $1\times1$, $1\times1$ convolutions).
In fig.~\ref{fig:resnet_attention_schematics} we provide a schematic illustration of  the different depth case for residual network architectures. 

Without loss of generality, we assume that transfer losses are placed between student and teacher attention maps of  same spatial resolution, but, if needed, 
 attention maps can be interpolated to match their shapes.  Let $S$, $T$ and $\mathbf{W}_S$, $\mathbf{W}_T$ denote student, teacher and their weights correspondingly, and let $\mathcal{L}(\mathbf{W}, x)$ denote a standard cross entropy loss. Let also $\mathcal{I}$ denote the indices of all teacher-student activation layer pairs for which we want to transfer attention maps. Then  we can define the following total loss:

 \vspace{-0.4cm}
\begin{equation}
  \mathcal{L}_{AT} =  \mathcal{L}(\mathbf{W}_S, x) +\frac{\beta}{2}\sum_{j\in \mathcal{I}}\| \frac{Q_S^j}{\|Q_S^j\|_2} - \frac{Q_T^j}{\|Q_T^j\|_2} \|_p\enspace,
  \label{eq:single_at}
\end{equation}

where $Q^j_S = vec(F(A_S^j))$ and $Q^j_T = vec(F(A_T^j))$ are respectively the $j$-th pair of student and teacher attention maps in vectorized form, and $p$ refers to norm type (in the experiments we use $p=2$).  As can be seen, during attention transfer we make use of $l_2$-normalized attention maps, i.e., we replace each vectorized attention map $Q$ with $ \frac{Q}{\|Q\|_2}$ ($l_1$ normalization could be used as well). It is worth emphasizing  that normalization  of attention maps is important for the success of the student  training.

Attention transfer can also be  combined with knowledge distillation \cite{KD}, in which case an additional term (corresponding to the cross entropy between  softened distributions  over labels of teacher and student)  simply needs to be included to the above loss.
 When combined, attention transfer adds very little computational cost, as attention maps for teacher can be easily computed during forward propagation, needed for distillation.

\subsection{Gradient-based attention transfer}

In this case we define attention as gradient w.r.t.\ input, which can be viewed as an input sensitivity map \cite{simonyan14}, i.e., attention at an input spatial location  encodes how sensitive the output prediction is w.r.t.  changes at that input location (e.g., if small changes at a pixel can have a large effect on the network output then it is logical to assume that the network is ``paying attention'' to that pixel).   Let's define the gradient of the loss w.r.t\ input for teacher and student as:

\vspace{-0.2cm}
\begin{equation}
  J_S = \frac{\partial}{\partial x} \mathcal{L}(\mathbf{W_S}, x), J_T = \frac{\partial}{\partial x} \mathcal{L}(\mathbf{W_T}, x)
\end{equation}

Then if we want student gradient attention to be similar to teacher attention, we can minimize a distance between them (here we  use $l_2$ distance but other distances can be employed as well):

\vspace{-0.2cm}
\begin{equation}
  \mathcal{L}_{AT}(\mathbf{W_S},\mathbf{W_T},x) = \mathcal{L}(\mathbf{W_S}, x) + \frac{\beta}{2}||J_S - J_T||_2
\end{equation}

As $\mathbf{W}_T$ and $x$ are given, to get the needed derivative w.r.t.\ $\mathbf{W}_S$:

\vspace{-0.2cm}
\begin{equation}
  \frac{\partial}{\partial \mathbf{W_S}} \mathcal{L}_{AT} =
  \frac{\partial}{\partial \mathbf{W_S}} \mathcal{L}(\mathbf{W_S}, x) + \beta(J_S - J_T) \frac{\partial^2}{\partial \mathbf{W_S}\partial x}\mathcal{L}(\mathbf{W_S}, x)
\end{equation}

So to do an update we first need to do forward and back propagation to get $J_S$ and $J_T$, compute the second error $\frac{\beta}{2}||J_S - J_T||_2$ and propagate it second time. The second propagation is similar to forward propagation in this case, and involves second order mixed partial derivative calculation $\frac{\partial^2}{\partial W_S\partial x}$. The above computation is similar to the double backpropagation technique developed by \cite{drucker-lecun-92} (where the $l_2$ norm of the gradient w.r.t.  input is used as  regularizer). Furthermore, it can be implemented efficiently in a framework with automatic differentiation support, even for modern architectures with sophisticated graphs. The second backpropagation has approximately  the same cost with first backpropagation, excluding forward propagation. 

We also propose to enforce horizontal flip invariance on gradient attention maps. To do that we propagate horizontally flipped images as well as originals, backpropagate and flip gradient attention maps back. We then add $l_2$ losses on the obtained attentions and outputs, and do second backpropagation:

\vspace{-0.4cm}
\begin{equation}\label{eq:flip_inv}
  \mathcal{L}_{sym}(\mathbf{W},x) = \mathcal{L}(\mathbf{W}, x) +
  \frac{\beta}{2}||\frac{\partial}{\partial x} \mathcal{L} (\mathbf{W}, x) - \mathrm{flip}(\frac{\partial}{\partial x} \mathcal{L} (\mathbf{W}, \mathrm{flip}(x))) ||_2\enspace,
\end{equation}

where $\mathrm{flip}(x)$ denotes the flip operator. This is similar to Group Equivariant CNN approach by \cite{DBLP:journals/corr/CohenW16}, however it is not a hard constraint. We experimentally find that this has a regularization effect on training.

We should note that in this work we consider only gradients w.r.t. the input layer, but in general one might  have the proposed attention transfer and symmetry constraints w.r.t. higher layers of the network.

\section{Experimental section}\label{sec:results}

In the following section we explore attention transfer on various image classification datasets. We split the section in two parts, in the first we include activation-based attention transfer and gradient-based attention transfer experiments on CIFAR, and in the second activation-based  attention transfer experiments on larger datasets.
For activation-based attention transfer we used Network-In-Network \cite{nin} and ResNet-based architectures (including the recently introduced Wide Residual Networks (WRN) \cite{Zagoruyko2016WRN}), as they are most performant and set strong baselines in terms of number of parameters compared to AlexNet or VGG, and have been explored in various papers across small and large datasets. On Scenes, CUB and ImageNet we experimented with ResNet-18 and ResNet-34.
As for gradient-based attention, we constrained ourselves to Network-In-Network without batch normalization and CIFAR dataset, due to the need of complex automatic differentiation.

\subsection{CIFAR experiments}

We start with CIFAR dataset which has small $32\times32$ images, and after downsampling top activations have even smaller resolution, so there is not much space for attention transfer. Interestingly, even under this adversarial setting, we find that  attention transfer seems to give reasonable benefits, offering in all cases consistent improvements. We use horizontal flips and random crops data augmentations, and all networks have batch normalization. We find that ZCA whitening has negative effect on validation accuracy, and omit it in favor of simpler meanstd normalization. We raise Knowledge Distillation (KD) temperature for ResNet transfers to 4, and use  $\alpha = 0.9$ (see \cite{KD} for an explanation of these parameters).

\subsubsection{Activation-based attention transfer}

Results of attention transfer (using  $F^2_{\mathrm{sum}}$ attention maps) for various networks on CIFAR-10 can be found in table \ref{table:at_cifar}. We experimented with teacher/student having the same depth (WRN-16-2/WRN-16-1), as well as different depth (WRN-40-1/WRN-16-1, WRN-40-2/WRN-16-2). In all combinations, attention transfer (AT) shows significant improvements, which are also higher when it is combined with knowledge distillation (AT+KD).

\begin{table*}[ht]
  \centering\small
  \begin{tabular}{|c|c|c|c|c|c|c|c|}
    \hline
    student & teacher & student & AT & F-ActT & KD & AT+KD & teacher  \\ \hline
    NIN-thin, 0.2M & NIN-wide, 1M   & 9.38 & 8.93 & 9.05 & 8.55 & 8.33 & 7.28 \\
    WRN-16-1, 0.2M & WRN-16-2, 0.7M & 8.77 & 7.93 & 8.51 & 7.41 & 7.51 & 6.31 \\
    WRN-16-1, 0.2M & WRN-40-1, 0.6M & 8.77 & 8.25 & 8.62 & 8.39 & 8.01 & 6.58 \\
    WRN-16-2, 0.7M & WRN-40-2, 2.2M & 6.31 & 5.85 & 6.24 & 6.08 & 5.71 & 5.23 \\ \hline
  \end{tabular}
  \caption{Activation-based attention transfer (AT) with various architectures on CIFAR-10. Error is computed as median of 5 runs with different seed. F-ActT means full-activation transfer (see~\S\ref{sec:F-ActT}).}
  \label{table:at_cifar}
\end{table*}

To verify if having at least one activation-based attention transfer loss per group in WRN transfer is important, we trained three networks with only one  transfer loss per network in \texttt{group1}, \texttt{group2} and \texttt{group3} separately, and compared to a network trained with all three losses. The corresponding results were 8.11, 7.96, 7.97 (for the separate losses) and  7.93 for the combined loss (using WRN-16-2/WRN-16-1  as teacher/student pair). Each loss provides some additional degree of attention transfer.

We also explore which attention mapping functions tend to work best using WRN-16-1 and WRN-16-2 as student and teacher networks respectively (table~\ref{table:cifar_at_functions}). Interestingly, sum-based functions work very similar, and better than max-based ones. From now on, we  will use sum of squared attention mapping function $F_{\mathrm{sum}}^2$ for simplicity. As for parameter $\beta$ in eq.~\ref{eq:single_at}, it usually varies about 0.1, as we set it to $10^3$ divided by number of elements in attention map and batch size for each layer. In case of combinining AT with KD we decay it during traning in order to simplify learning harder examples.

\subsubsection{Activation-based AT vs. transferring full activation}\label{sec:F-ActT}

To check if transferring information from full activation tensors is more beneficial than from attention maps, we experimented with FitNets-style hints using $l_2$ losses on full activations directly, with $1\times1$ convolutional layers to match tensor shapes, and found that improvements over baseline student were minimal (see column F-ActT in table~\ref{table:at_cifar}). For networks of the same width different depth we tried to regress directly to activations, without $1\times1$ convolutions. We also use $l_2$ normalization before transfer losses, and decay $\beta$ in eq.~\ref{eq:single_at} during training as these give better performance. We find that AT, as well as full-activation transfer, greatly speeds up convergence, but AT gives much better final accuracy improvement than full-activation transfer (see fig.~\ref{fig:wrn_at_fitnet}, Appendix). It seems quite interesting that attention maps  carry information that is more important for transfer than full activations.

\begin{minipage}{0.33\textwidth}
  \centering\small
  \begin{tabular}{|l|c|}
    \hline
    attention mapping function & error \\ \hline
    no attention transfer   & 8.77 \\
    $F_{\mathrm{sum}}$  & 7.99 \\
    $F^2_{\mathrm{sum}}$ & 7.93\\
    $F^4_{\mathrm{sum}}$ & 8.09 \\
    $F^1_{\mathrm{max}}$ & 8.08 \\
    
    \hline
  \end{tabular}
  \captionof{table}{Test error of WRN-16-2/WRN-16-1 teacher/student pair for various attention mapping functions. Median of 5 runs test errors are reported.}
  \label{table:cifar_at_functions}
\end{minipage}
\hspace{0.03\textwidth}
\begin{minipage}{0.67\textwidth}
  \centering\small
  \begin{tabular}{|l|c|}
    \hline
    norm type & error \\
    \hline
    baseline (no attention transfer) & 13.5 \\
    min-$l_2$ \cite{drucker-lecun-92} & 12.5 \\
    grad-based AT & 12.1 \\
    KD & 12.1 \\
    symmetry norm & 11.8 \\
    activation-based AT & \bf{11.2} \\
    \hline
  \end{tabular}
  \captionof{table}{Performance of various gradient-based attention methods on CIFAR-10. Baseline is a thin NIN network with 0.2M parameters (trained only on horizontally flipped augmented data and without batch normalization), min-$l_2$ refers to using $l_2$ norm of gradient w.r.t.\ input as regularizer, symmetry norm - to using flip invariance on gradient attention maps (see eq. \ref{eq:flip_inv}), AT - to attention transfer, and KD - to Knowledge Distillation (both\ AT\ and KD use a wide NIN of 1M parameters as teacher).}
  \label{table:grad-based_cifar}
\end{minipage}

\subsubsection{Gradient-based attention transfer}

For simplicity we use thin Network-In-Network model in these experiments, and don't apply random crop data augmentation with batch normalization, just horizontal flips augmentation. We also only use deterministic algorithms and sampling with fixed seed, so reported numbers are for single run experiments. We find that in this setting network struggles to fit into training data already, and turn off weight decay even for baseline experiments. In future we plan to explore gradient-based attention for teacher-student pairs that make use of batch normalization, because it is so far unclear how batch normalization should behave in the second backpropagation step required during gradient-based attention transfer (e.g., should it contribute to batch normalization parameters, or is a separate forward propagation with fixed parameters needed).

We explored the following methods:

\begin{itemize}
  \item Minimizing $l_2$ norm of gradient  w.r.t. input, i.e.\ the double backpropagation method \cite{drucker-lecun-92};
  \item Symmetry norm on gradient attention maps (see eq.~\ref{eq:flip_inv});
  \item Student-teacher gradient-based attention transfer;
  \item Student-teacher activation-based attention transfer.
\end{itemize}

Results for various methods are shown in table~\ref{table:grad-based_cifar}. Interestingly, just minimizing $l_2$ norm of gradient already works pretty well. Also, symmetry norm is one the best performing attention norms, which we plan to investigate in future on other datasets as well. We also observe that, similar to activation-based attention transfer, using gradient-based attention transfer leads to improved performance. We also trained a network with activation-based AT in the same training conditions, which resulted in the best performance among all methods. We should note that the architecture of student NIN without batch normalization is slightly different from teacher network, it doesn't have ReLU activations before pooling layers, which leads to better performance without batch normalization, and worse with. So to achieve the best performance with activation-based AT we had to train a new teacher, with batch normalization and without ReLU activations before pooling layers, and have AT losses on outputs of convolutional layers.

\subsection{Large input image networks}

In this section we experiment with hidden activation attention transfer on ImageNet networks which have $224\times224$ input image size. Presumably, attention matters more in this kind of networks as spatial resolution of attention maps is higher.

\subsubsection{Transfer learning}

To see how attention transfer works in finetuning we choose two datasets: Caltech-UCSD Birds-200-2011 fine-grained classification (“CUB”) by \cite{WahCUB_200_2011}, and MIT indoor scene classification (“Scenes”) by \cite{scenes}, both containing around 5K images training images. We took ResNet-18 and ResNet-34 pretrained on ImageNet and finetuned on both datasets. On CUB we crop bounding boxes, rescale to 256 in one dimension and then take a random crop. Batch normalization layers are fixed for finetuning, and first group of residual blocks is frozen. We then took finetuned ResNet-34 networks and used them as teachers for ResNet-18 pretrained on ImageNet, with $F^2_{\mathrm{sum}}$ attention losses on 2 last groups. In both cases attention transfer provides significant improvements, closing the gap between ResNet-18 and ResNet-34 in accuracy. On Scenes AT works as well as KD, and on CUB AT works much better, which we speculate is due to importance of intermediate attention for fine-grained recognition. Moreover, after finetuning, student's attention maps indeed look more similar to teacher's (Fig.~\ref{figure:scenes_sumabs_maps}, Appendix).

\begin{table}
  \vspace{-0.1cm}
  \centering\small
  \begin{tabular}{|c|c|c|c|}
    \hline
    type & model & ImageNet$\rightarrow$CUB & ImageNet$\rightarrow$Scenes \\ \hline
    student & ResNet-18 & 28.5 & 28.2 \\
    KD      & ResNet-18 & 27 (-1.5)   & 28.1 (-0.1) \\
    AT      & ResNet-18 & 27 (-1.5)   & 27.1 (-1.1) \\
    teacher & ResNet-34 & 26.5 & 26 \\ \hline
  \end{tabular}
  \caption{Finetuning with attention transfer error on Scenes and CUB datasets}
  \label{at:scenes_cub}
\end{table}

\subsubsection{ImageNet}

To showcase activation-based attention transfer on ImageNet we took ResNet-18 as a student, and ResNet-34 as a teacher, and tried to improve ResNet-18 accuracy. We added only two losses in the 2 last groups of residual blocks and used squared sum attention $F^2_{\mathrm{sum}}$. We also did not have time to tune any hyperparameters and kept them from finetuning experiments. Nevertheless, ResNet-18 with attention transfer achieved 1.1\% top-1 and 0.8\% top-5 better validation accuracy (Table.~\ref{table:at_imagenet_top5} and Fig.~\ref{fig:at_imagenet_top5}, Appendix), we plan to update the paper with losses on all 4 groups of residual blocks.

We were not able to achieve positive results with KD on ImageNet. With ResNet-18-ResNet-34 student-teacher pair it actually hurts convergence with the same hyperparameters as on CIFAR. As it was reported that KD struggles to work if teacher and student have different architecture/depth (we observe the same on CIFAR), so we tried using the same architecture and depth for attention transfer. On CIFAR both AT and KD work well in this case and improve convergence and final accuracy, on ImageNet though KD converges significantly slower (we did not train until the end due to lack of computational resources). We also could not find applications of FitNets, KD or similar methods on ImageNet in the literature. Given that, we can assume that proposed activation-based AT is the first knowledge transfer method to be successfully applied on ImageNet.

\section{Conclusions}\label{sec:conclusions}

We presented several ways of transferring attention from one network to another, with experimental results over several image recognition datasets. It would be interesting to see how attention transfer works in cases where spatial information is more important, e.g.\ object detection or weakly-supervised localization, which is something that we plan to explore in the future.

Overall, we think that our interesting findings will help further advance knowledge distillation, and understanding convolutional neural networks in general.

\bibliography{iclr2017_conference}
\bibliographystyle{iclr2017_conference}

\newpage
\appendix
\section{Appendix}
\subsection{Figures and tables}

\begin{figure}[ht]
  \centering
  \includegraphics[scale=0.46]{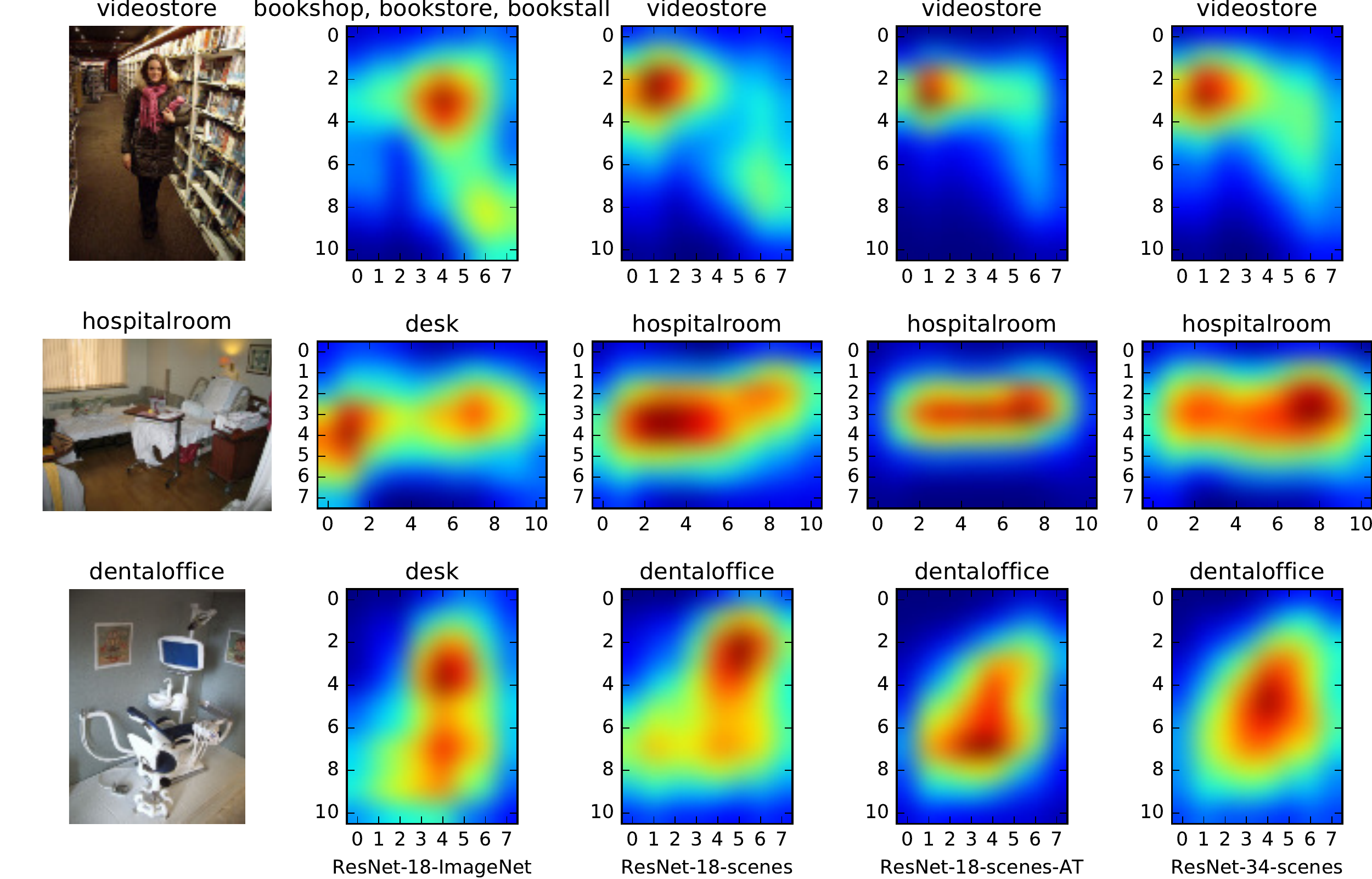}
  \caption{Top activation attention maps for different Scenes networks: original pretrained ResNet-18 (ResNet-18-ImageNet), ResNet-18 trained on Scenes (ResNet-18-scenes), ResNet-18 trained with attention transfer (ResNet-18-scenes-AT) with ResNet-34 as a teacher, ResNet-34 trained on Scenes (ResNet-34-scenes). Predicted classes for each task are shown on top. Attention maps look more similar after transfer (images taken from test set).}
  \label{figure:scenes_sumabs_maps}
\end{figure}

\def\figatscale{0.42}
\begin{figure}[ht]
  \centering\small
  \subfigure[Attention transfer on ImageNet between ResNet-18 and ResNet-34. Solid lines represent top-5 validation error, dashed - top-5 training error. Two attention transfer losses were used on the outputs of two last groups of residual blocks respectively, no KD losses used.]{
    \includegraphics[scale=\figatscale]{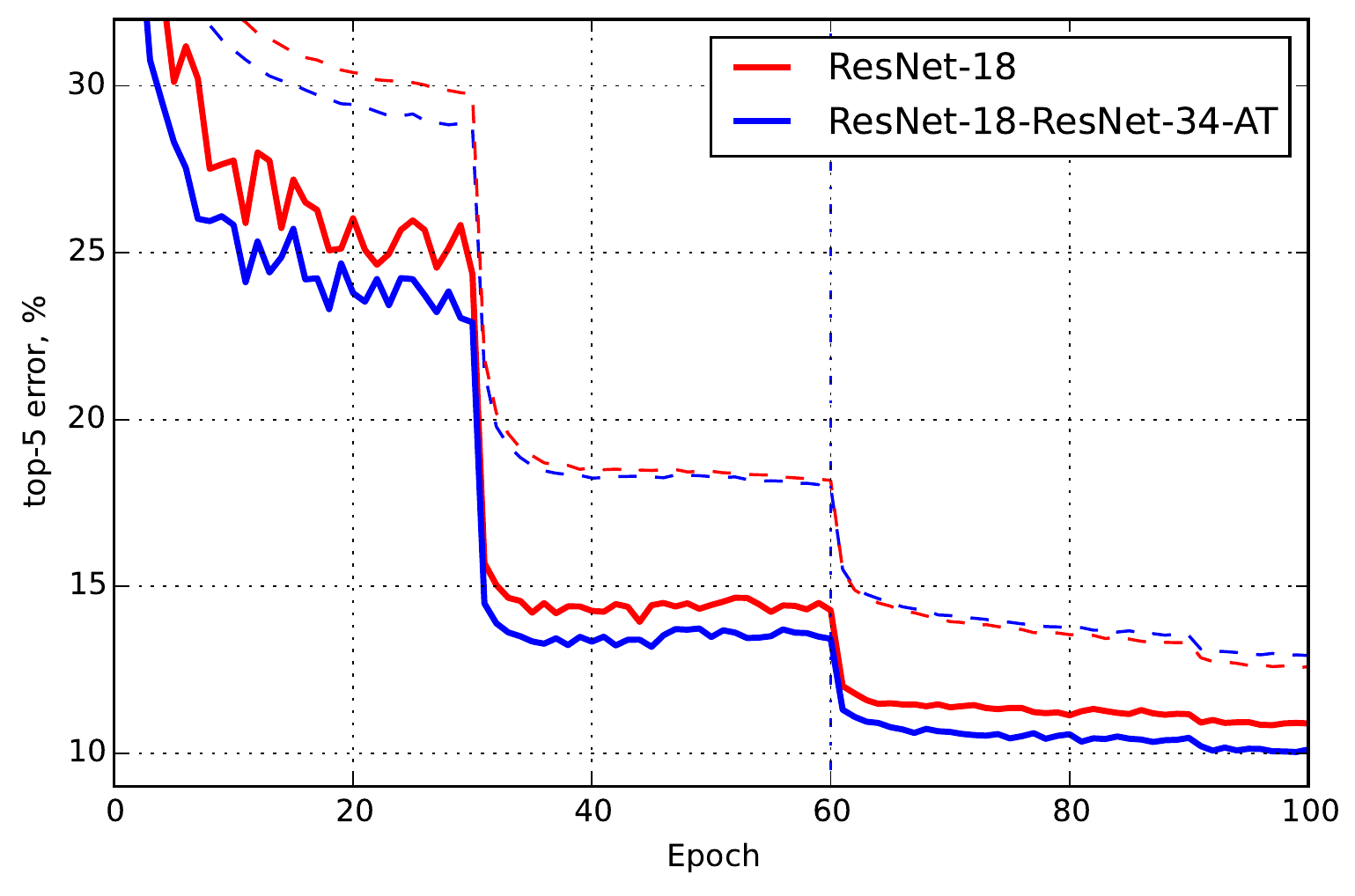}
    \label{fig:at_imagenet_top5}
  }
  \hspace{0.2cm}
  \subfigure[Activation attention transfer on CIFAR-10 from WRN-16-2 to WRN-16-1. Test error is in bold, train error is in dashed lines. Attention transfer greatly speeds up convergence and improves final accuracy.]{
    \includegraphics[scale=\figatscale]{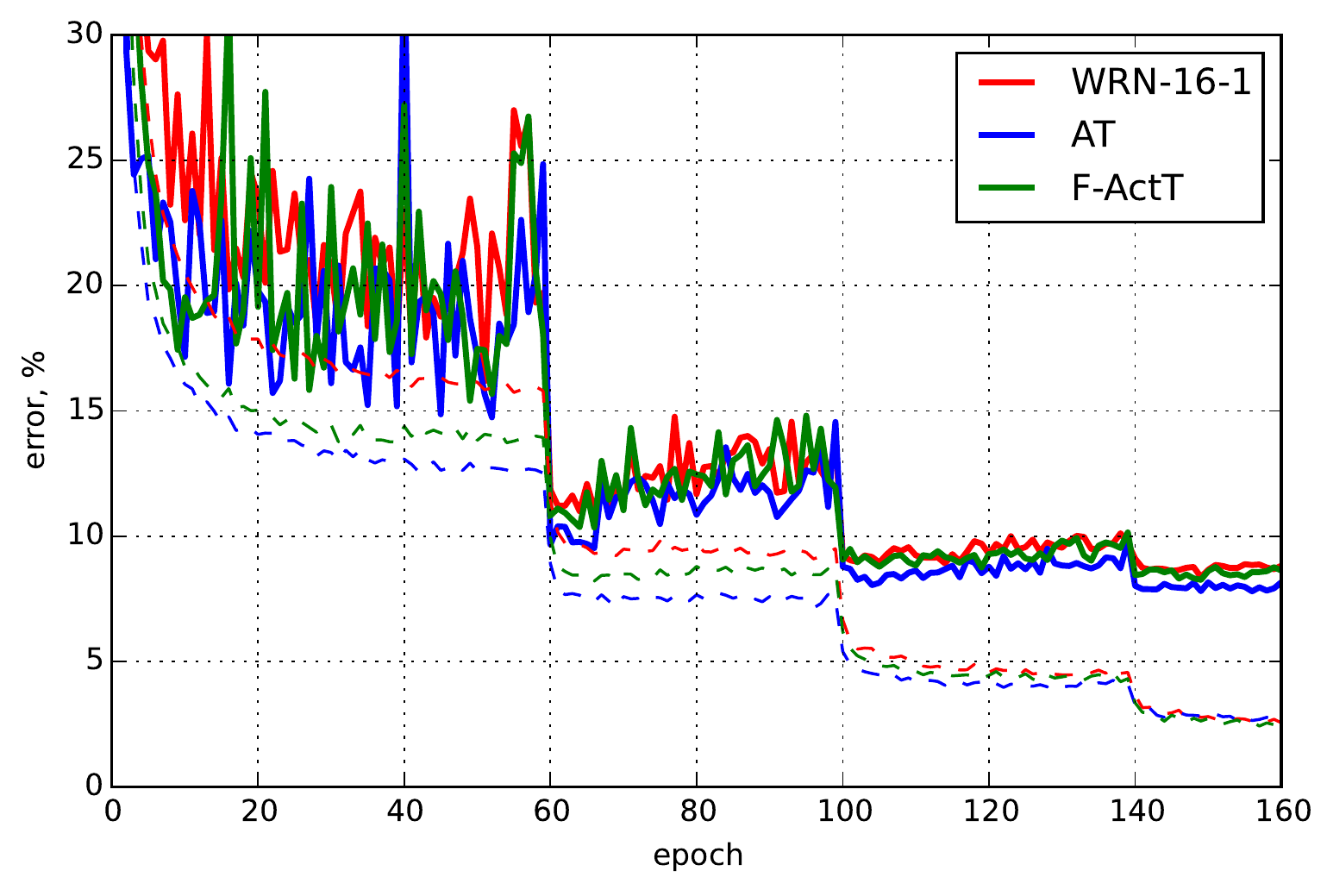}
    \label{fig:wrn_at_fitnet}
  }
  \vspace{-0.2cm}
  \caption{}
\end{figure}

\begin{table}[h]
  \centering\small
  \begin{tabular}{|l|c|}
    \hline
    Model & top1, top5 \\ \hline
    ResNet-18 & 30.4, 10.8 \\
    AT & 29.3, 10.0 \\
    ResNet-34 & 26.1, 8.3 \\ \hline
  \end{tabular}
  \caption{Attention transfer validation error (single crop) on ImageNet. Transfer losses are added on epoch 60/100.}
  \label{table:at_imagenet_top5}
\end{table}

\newpage
\subsection{Implementation details}

The experiments were conducted in Torch machine learning framework. Double propagation can be implemented in a modern framework with automatic differentiation support, e.g.\ Torch, Theano, Tensorflow. For ImageNet experiments we used fb.resnet.torch code, and used 2 Titan X cards with data parallelizm in both teacher and student to speed up training. Code and models for our experiments are available at \url{https://github.com/szagoruyko/attention-transfer}.

\end{document}